\documentclass{article}

\PassOptionsToPackage{numbers, sort}{natbib}

     \usepackage[preprint]{d3s3_neurips_2024}

\usepackage{graphicx}
\usepackage{subfigure}
\usepackage{float}
\usepackage{wrapfig}

\usepackage[utf8]{inputenc} 
\usepackage[T1]{fontenc}    
\usepackage{hyperref}       
\usepackage{url}            
\usepackage{booktabs}       
\usepackage{amsfonts}       
\usepackage{nicefrac}       
\usepackage{microtype}      
\usepackage{xcolor}         
\usepackage{acro}
\usepackage{amsmath}
\usepackage{amssymb}
\usepackage{mathtools}
\usepackage{amsthm}
\usepackage{bbm}

\usepackage[utf8]{inputenc}
\usepackage[T1]{fontenc}

\usepackage[disable,textsize=tiny]{todonotes}

\title{Spatial Shortcuts in Graph Neural Controlled Differential Equations}

\author{%
	Michael Detzel\thanks{Corresponding author michael.detzel@hhi.fraunhofer.de} \\
	Fraunhofer HHI\\
	\And
	Gabriel Nobis \\
	Fraunhofer HHI\\
	\AND
	Jackie Ma \\
	Fraunhofer HHI\\
	\And
	Wojciech Samek \\
	Fraunhofer HHI and TU Berlin\\
}

\DeclareAcronym{rnn}{
	short=RNN,
	long=Recurrent Neural Network}
\DeclareAcronym{nde}{
	short=NDE,
	long=Neural Differential Equation}
\DeclareAcronym{gncde}{
	short=GNCDE,
	long=Graph Neural Controlled Differential Equation}
\DeclareAcronym{ncde}{
	short=NCDE,
	long=Neural Controlled Differential Equation}
\DeclareAcronym{nn}{
	short=NN,
	long=Neural Network}
\DeclareAcronym{node}{
	short=NODE,
	long=Neural Ordinary Differential Equation}
\DeclareAcronym{pde}{
	short=PDE,
	long=Partial Differential Equation}
\DeclareAcronym{gnn}{
	short=GNN,
	long=Graph Neural Network}
\DeclareAcronym{lstm}{
	short=LSTM,
	long=Long Short Term Memory}
\DeclareAcronym{gru}{
	short=GRU,
	long=Gated Recurrent Unit}
\DeclareAcronym{agc}{
	short=AGC,
	long=Adaptive Graph Convolution
	}
\DeclareAcronym{resnet}{
	short=ResNet,
	long=Residual Neural Network}
\DeclareAcronym{mae}{
	short=MAE,
	long=Mean Absolute Error}

\newcommand*\dif{\mathop{}\!\mathrm{d}}

\begin{document}

	\maketitle
		
	\begin{abstract}
	 We incorporate prior graph topology information into a \ac{ncde} to predict the future states of a dynamical system defined on a graph. The informed \ac{ncde} infers the future dynamics at the vertices of simulated advection data on graph edges with a known causal graph, observed only at vertices during training. We investigate different positions in the model architecture to inform the NCDE with graph information and identify an outer position between hidden state and control as theoretically and empirically favorable. Our such informed \ac{ncde} requires fewer parameters to reach a lower \ac{mae} compared to previous methods that do not incorporate additional graph topology information.
	\end{abstract}
	\section{Introduction}
	Effect follows cause. When a problem is represented on a graph, its structure contains information on how causes at one spatial position are linked to effects at another. To learn about physical dynamics from spatial time series data, one can often leverage this structural information. \citet{scholkopf_causality_2022b} describes differential equations as the gold standard for understanding cause-effect structures and highlight the lack of a time component in statistical machine learning methods. \acp{nde} \citep{chen_neural_2019} are able to learn a hidden state that evolves continuously in time, and could remedy this lack. With Neural Controlled Differential Equations (NCDE) \citep{kidger_neural_2020}, one can update the hidden state continuously with data incoming at different points in time. If one also wants to account for spatial dependencies, \acp{gncde} \citep{choi_graph_2021} can be used, where a node embedding is learned to capture these spatial dependencies.
We incorporate prior known graph topology information into a \ac{gncde} to infer the future dynamics at the vertices. We therefore generate data coming from a graph advection simulation from which we know the underlying graph topology plus the temporal cause and effect relation. We will outline the close connection between the graph information in the data generated and the artificial \ac{nn} architecture. Then we train our Informed \acp{gncde} and let them learn the dynamics to predict their future behavior. We start by describing advection on graphs and the theory before we explain the data generation and our Informed \ac{gncde}. We believe this approach can lead to improvements in domains where graph information is available or where time series data is scarce or partially missing. \acp{ncde} are an effective method in this context  due to their ability to handle irregular time series data \citep{kidger_neural_2022}. With graph information one has the potential to predict with fewer observations, as the graph structure itself does not have to be learned on top of the temporal dynamics. Promising domains for introducing known graph structure are traffic forecasting, river water level forecasting, climate and weather prediction, or disease spread (see \ref{applications}). 
	\begin{figure*}[t]
		\begin{tabular}{l l l}
			\includegraphics[width=0.24\linewidth]{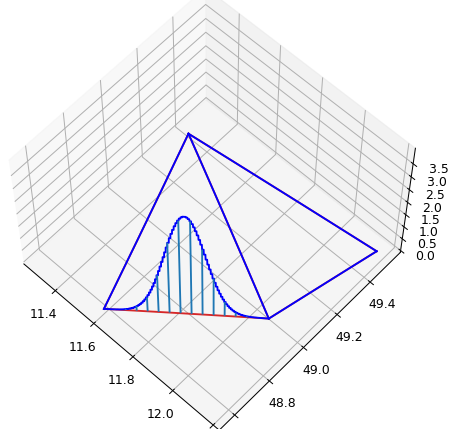}
			\includegraphics[width=0.24\linewidth]{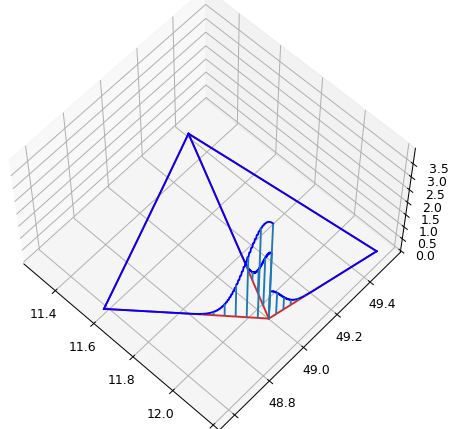} 
			\includegraphics[width=0.24\linewidth]{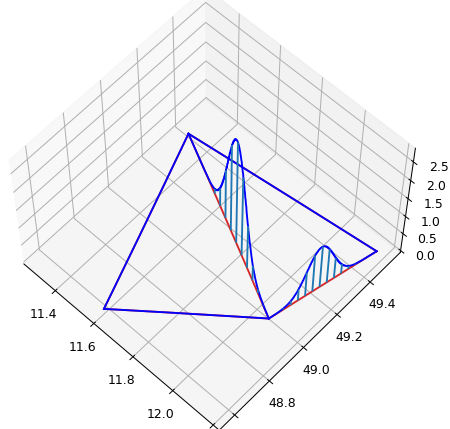}  
			\includegraphics[width=0.24\linewidth]{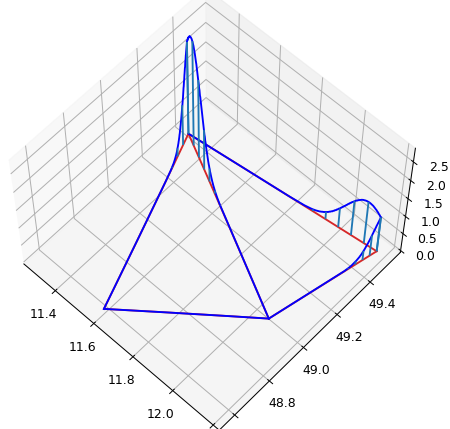} 
		\end{tabular}
		\caption{Advection of an initial Gaussian pulse on a graph with 5 edges over time}\label{simulation_visual}
	\end{figure*}
	
	\section{Related Work}
	\citet{choi_graph_2021} consider time series on a graph by using a graph embedding to capture the spatial dependencies, without directly incorporating prior graph knowledge for traffic forecasting.
	Spatial-Temporal \acp{gnn} \citep{wu_comprehensive_2021,rossi2020temporal} inform \acp{nn} with graph information, where Message Passing \acp{nn} \cite{gilmer_2017_message_passing} are combined with \acp{rnn} like \acp{lstm} \citep{hochreiter1997long} or \acp{gru} \citep{cho2014learning}.
	\citet{poli_graph_2021} introduce spatial inductive bias as a node embedding into the vector field while generalizing the notion of \acp{node} to graphs. 
	Other attempts of incorporating prior knowledge into \ac{nn} architecture include Physics Informed Neural Networks (PINNS) \citep{raissi_physics-informed_2019} which enforce a prior defined \ac{pde} but do not act on graphs as they avail oneself of a discretized continuous description of the spatial domain. \citet{kosma2023neural} use \acp{node} for epidemic spreading on graphs and \citet{verma2024climode} use \acp{node} for climate and weather forecasting in discretized continuous space.
	\section{Learning Advection with Graph Neural Controlled Differential Equations}
	\textbf{Advection on Graph Edges.} In contrast to \citet{chapman2011advection}, where the quantities on the vertices are updated discretely in time, we advect quantities that move continuously in time along the edges of a graph and fork or merge at vertices. 
	For the advection simulation, we consider a directed graph $\mathcal{G}$ with a set of vertices $\mathcal{V}$ and edges $\mathcal{E}$. Every edge $e \in \mathcal{E}$ has a time-static continuous domain $\Omega_e = \{x  \in \mathbb{R} $ $|$ $ 0 \leq x < \delta(e)\}$ and a function space $V_{e} = \{y_{e}(x,t)\vert\Omega(e) \times \mathcal{T} \rightarrow \mathbb{R}^d \}$ attached to it, where $\delta(e)$ is the length of edge $e$ and $t \in \mathcal{T} = [0,T]$ is a point in time. We first focus on the description of the advection on the interior of the edges without considering transitions between edges:
	Given an initial state of the dynamic system it evolves according to an advection differential equation
	\begin{equation}\label{advection0}
		\frac{\partial \bold{y}}{\partial t}+\nabla \cdot(\bold{y} \odot \mathbf{v}) = 0, \quad \bold{y}(\ \cdot \ , 0) =  \bold{y}_0	
	\end{equation}
	with $\bold{v} = (v_1, \ \ldots,\ v_{\vert \mathcal{E} \vert})^\intercal$ being the vector of scalar velocity fields $v_e: \Omega_e \rightarrow \mathbb{R}$, with $e \in \{1,\ldots, \vert \mathcal{E} \vert\}$, that advect the quantity $\bold{y} = (y_{e_1},\ \ldots,\ y_{e_{\vert \mathcal{E} \vert}})^\intercal$ through space over time. The divergence is intended row-wise and $\odot$ signifies element-wise multiplication.
	We restrict the entries of $\bold{y}$ to positive values, so that the flow direction is always the direction of the directed edge.
	We assume that the velocity field is constant. In the case of a one-dimensional quantity and velocity field along the edges, this leads to $ \frac{\partial y_e}{\partial t} = -{v_e} \frac{\partial y_e}{\partial x},$
	which has the  analytical solution $y_e(x,t) = y^0_e(x-vt)$, with initial condition $y^0_e := y_e(x,0)$, for all $x \in \Omega(e)$, for all $e \in \mathcal{E}$ and $v_e t \leq x \leq \delta(e)$. With this description of the dynamic, $y_e$ for $e \in \mathcal{E}$ is not required to be differentiable. Now we describe the dynamic on the transition between adjacent edges. For $x < v_e t$ one would look back "beyond the edge". We therefore resort to a matrix formulation to represent the edge transitions of $y$: 
	\begin{equation}\label{edge_transition_eq}
		\bold{y}(\bold{x}, t) = \big( \bold{A}_{\mathcal{E}} \bold{y}(\bold{x}, 0) \big) \Big |_{\bold{x} = (\delta(\bold{e}) - v_e t \  \bold{1}_{|\mathcal{E}|})}, \quad x<v_e t,
	\end{equation}
	where $t < \min_{e \in \mathcal{E}} \delta(e) /v_e$ has to hold for every edge to not look further back than the previous edge and
	where $\bold{A}_{\mathcal{E}}$ is a directed "edge transition matrix", signifying to which edge a quantity transitions when the quantity reaches the end of an edge. $\bold{A}_{\mathcal{E}}$  acts as a function operator on the vector of functions $\bold{y}_0 = (y^0_{1},\ \ldots,\  y^0_{|\mathcal{E}|})^\intercal$, and $\bold{x} =(x_{1},\ \ldots, \ x_{|\mathcal{E}|})^\intercal$ is a  collection vector of spatial coordinates on the different edges and $\bold{1} = (1,\ \ldots ,\ 1)^\intercal \in \mathbb{R}^{|\mathcal{E}|}$. The entries $a_{ij}$ of $\bold{A}_{\mathcal{E}}$ are defined as
	\begin{align*}
		a_{ij} =
		\begin{cases}
			p_{ij}  \in (0,1], &\text{ if } e_i \text{ follows on } e_j \\
			0, &\text{ else}
		\end{cases},
	\end{align*} with $p_{ij}$ being the proportion of the quantity on $e_i$ transported to $e_j$ and $\sum_{i \in  \mathcal{E}} p_{ij} = 1$ to enforce conservation. 
	The matrix $\bold{A}_{\mathcal{E}}$ can be obtained from the the vertex adjacency matrix $\bold{A}_{\mathcal{V}}$ (details in \ref{edge_transition}). For the numerical simulation the graph edges are spatially discretized into 100 segments $s_k$ from which $50$ are uniformly randomly selected without replacement to be initialized with discrete uniformly sampled values $\psi_{e,s_k} \sim \mathcal{U}\{0,10\}, k \in \{1,100\}$. We then iteratively advect with a recurrence equation derived from Equation (\ref{edge_transition_eq}) to derive
		\begin{equation*}
		\bold{y}(\bold{x}, t + \Delta t ) = \big( \bold{A}_{\mathcal{E}} (\bold{y}(\bold{x}, t)) \big) \Big |_{\bold{x} = (\delta(\bold{e}) - v \Delta t \ \bold{1}_{|\mathcal{E}|})}
		\end{equation*}
	and use $\Delta t = 5$ minutes to evolve for 48 steps into the future. To obtain data measurements at the nodes, the quantity that passed through the vertex in a given time span is aggregated. We take measurements for $\Delta t$ and acquire time series on the graph nodes that are used for the \ac{nn} training. As input data for the \acp{gncde}, we are facing  a time-series of graphs $\{ \mathcal{G}_{t_i} \triangleq (\mathcal{V}, \mathcal{E}, \bold{Y_{t_i}})\}_{i=0}^{N}$, with $\bold{Y_{t_i}}$ being the aggregated quantities at the vertices. 

	\textbf{Incorporating topology knowledge into \acp{gncde}.}
	Let us review the notion of \acp{ncde} \citep{kidger_neural_2020}: A \ac{ncde} is described by 
	\begin{equation*}
		\boldsymbol{z}(T)=\boldsymbol{z}(0)+\int_0^T f_{\boldsymbol{\theta}}\Big(\boldsymbol{z}(t) \Big) d\boldsymbol{x}(t),
	\end{equation*}
	where $T,t \in \mathbb{R^+}$, $\boldsymbol{z}(t)\in \mathbb{R}^{d_z}$, $f_{\boldsymbol{\theta}}: \mathbb{R}^{d_z} \rightarrow \mathbb{R}^{d_z} \times \mathbb{R}^{d_x}$ is a vector field modeled by an \ac{nn} with trainable parameters $\boldsymbol{\theta} \in \mathbb{R}^{d_{\boldsymbol{\theta}}}$ and $\boldsymbol{x}(t) \in \mathbb{R}^{d_x}$ is the control path. This means that the hidden state $\boldsymbol{z}(t))$ is continuously updated over time $t$ from $0$ up to time $T$ by the vector field $f_{\boldsymbol{\theta}}$ and the control $\bold{x}$. This formulation allows inputting data 
	at arbitrary times $t$ via $\bold{x}(t)$. The path of state $\boldsymbol{z}(t)$ encodes the information gathered from time $0$ to $T$ and is used to make the final predictions $\{{\hat{\bold{Y}}_{t_i}}\}_{i = 1}^{M}$ with $T< t_i\leq T+ \tau$ at time $T$ for the next $M$ time steps up to time $T+\tau$. One can solve and backpropagate through the \ac{ncde} given a loss function $\mathcal{L}(\{{\hat{\bold{Y}}_{t_i}(\bold{x})}\}_{i = 1}^{M},\{{{\bold{Y}}_{t_i}}\}_{i = 1}^{M})$.
	\citet{jhin_attentive_2024} and \citet[p. 66]{kidger_neural_2022} showed that \acp{ncde} can be coupled to use the encoded hidden path of a first NCDE as a control path input to a second NCDE. Firstly, analogously to \citet{choi_graph_2021}, we update a $d_{h}$-dimensional hidden state $\boldsymbol{H}(t) \in \mathbb{R}^{\vert \mathcal{V} \vert \times d_{h}}$ continuously in time for every node separately to capture the temporal dependencies via 
			\begin{wrapfigure}{r}{0.4\textwidth}
		\begin{tabular}{c c}
			\quad &
			\includegraphics[scale=0.6]{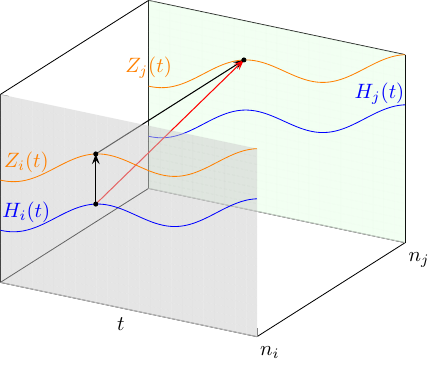}
		\end{tabular}
		\caption{Difference in information transport between hidden state $\bold{H}_{i}$ at node $n_i$ and hidden state $\bold{Z}_{j}$ at node $n_j$. Connection with $A_{\mathcal{V}}^{outer}$  leads to a direct connection {\color{red} (red)} versus connecting nodes via $A_{\mathcal{V}}^{inner}$ (black)}
		\label{3d_path}
	\end{wrapfigure}
	\begin{equation}\label{temporal}
		\boldsymbol{H}(T)=\boldsymbol{H}(0)+\int_0^T f_{\boldsymbol{\theta}}\Big(\boldsymbol{H}(t) \Big) \frac{d \boldsymbol{X}(t)}{d t} d t.
	\end{equation}
	Secondly, we utilize $\boldsymbol{H}$ as control to drive another \ac{ncde} to continuously update the hidden state $\boldsymbol{Z}(T) \in \mathbb{R}^{\vert \mathcal{V} \vert \times d_{z}}$, with $d_{z}$ being the dimension of the hidden state $\boldsymbol{Z}$ that contains now also spatial information via
	\begin{equation}\label{spatial}
		\boldsymbol{Z}(T)=\boldsymbol{Z}(0)+\int_0^T g_{\boldsymbol{\gamma}}\Big(\boldsymbol{Z}(t) \Big) \frac{d \boldsymbol{H}(t)}{d t} d t,
	\end{equation}
	where the vector field $g_{\boldsymbol{\gamma}}$ learns the spatial dependencies, making it a \ac{gncde}. We differentiate between the internal and external mappings that the vector field $g_{\boldsymbol{\gamma}}$ describes:
	\begin{align}
		g_{\boldsymbol{\gamma}}(\bold{Z}(t))&: \mathbb{R}^{d_z} \rightarrow \mathbb{R}^{d_z} \times \mathbb{R}^{d_h}, \label{inner_func}
		\\
		g_{\boldsymbol{\gamma}}(\bold{Z}(t))  \bullet \cdot &: \mathbb{R}^{d_h} \rightarrow \mathbb{R}^{d_z}  \label{mat_multi}, 
	\end{align}
	using $\bullet$ to emphasize the tensor product. Whereas Equation (\ref{inner_func}) describes the inner workings of the update of the hidden state $\bold{Z}$ on itself, the tensor multiplication in Equation (\ref{mat_multi}) describes the direct connection between control (and thereby the data) to the hidden state. Of course both happen simultaneously. 
	The function in Equation (\ref{inner_func}) describes how a hidden state entry at a node has to encode "where to look" based on the entries of other hidden states and itself, to then "find" the information in the control. 
	In contrast the function in Equation (\ref{mat_multi}) describes the situation in which one "knows where to look" (at least spatially) to read out the right hidden state update information from the control which is given. In Figure \ref{3d_path} differences in the connections are schematically depicted.		
	In our Informed \ac{gncde} architectures, we use exactly this second notion and introduce a vertex transition matrix $\bold{A}_{\mathcal{V}}^{outer}$ into the equation to obtain (\ref{dimensions} for dimensional details)
	\begin{equation}\label{outer_info}
	\boldsymbol{Z}(T)=\boldsymbol{Z}(0)+\int_0^T g_{\boldsymbol{\gamma}}\Big(\boldsymbol{Z}(t) \Big) \bold{A}_{\mathcal{V}}^{outer} \frac{d \boldsymbol{H}(t)}{d t} d t
	\end{equation}
	 with $\bold{A}_{\mathcal{V}}^{outer}$ corresponding to the edge transition matrix $\bold{A}_{\mathcal{E}}$. Alternatively, one can also introduce topology information into the \ac{gncde} inside $g_{\theta}$ by writing
	$ g_{\boldsymbol{\gamma}} = g_{3} \circ \bold{A}_{\mathcal{V}}^{inner} g_2 \circ g_1(\boldsymbol{Z}(t))$,
	where $g_l, \ l \in \{1,2,3\}$ are the inner \ac{nn} layers. \citet{choi_graph_2021} use a node embedding called \ac{agc} similar to the embedding used in \citet{kipf_semi-supervised_2017} inside $g_{\boldsymbol{\gamma}}$ to learn dependencies across graph nodes. The \ac{agc} is placed inside $g_2$ and $\bold{A}_{\mathcal{V}}^{inner}$ can be considered the identity matrix.
	One can set $\bold{A}_{\mathcal{V}}^{outer}$ to the identity matrix, as well, to obtain their \ac{gncde}, or one can set it to an undirected, directed or directed with prior known proportional weights vertex adjacency matrix. $\bold{A}_{\mathcal{V}}^{inner}$ can be chosen analogously.
	\section{Experiments}
	We sample 1000 and 10000 time-series for a 4-node graph and a 10-node graph (their adjacencies can be found in \ref{adjacencies}), respectively. $\bold{A}_{\mathcal{V}_4}$ represents a graph whose underlying topology is depicted in Figure \ref{simulation_visual}, $\bold{A}_{\mathcal{V}_{10}}$ represents a simple graph with subsequent edges, that exhibits strong sparsity. The task is to predict the quantities for the next 24 time points given the current and the previous 24 time points. We trained the Informed \ac{gncde} with different variations for the inner and outer graph mechanism and calculate the mean absolute error (MAE) on the test dataset. For the 4-node graph, the informedness on top of the AGC leads to a slight improvement in \ac{mae} compared to only using AGC of $2.67$ vs. $2.73$ (Table \ref{model_large}). Notably, using the outer mechanism alone leads to a \ac{mae} of $2.75$, with only requiring $293,184$ vs. $416,744$ parameters. For the 10-node graph with longer training time, we obtain lower \ac{mae} for just using the outer informedness versus only AGC and significant lower \ac{mae} for using informedness on top of AGC with \ac{mae} of $0.66$ vs. $1.42$ (Table \ref{model_large}).
	
	\begin{table}
		\begin{tabular}{cccccccccc}
			\toprule
			& \multicolumn{3}{c}{$\bold{A}_{\mathcal{V}^{outer}}$, $\vert \mathcal{V} \vert = 4$} & &  & \multicolumn{3}{c}{$\bold{A}_{\mathcal{V}^{outer}}, \vert \mathcal{V} \vert = 10$} & \\ \cmidrule{2-4} \cmidrule{7-9}
			& Identity & & Informed &  &  & Identity & & Informed \\ 
			$\bold{A}_{\mathcal{V}^{inner}}$ & MAE & & MAE & \# params & &  MAE & & MAE &\# params \\ \cmidrule{1-5} \cmidrule{7-10}
			Identity & 3.60 & & 2.75          &  293,184  & & 2.61 & & 1.11       &  293,244 \\ 
			Informed & 3.10 & & 3.52          &  293,184  & &  -   & & -          &  293,244 \\
			AGC 	 & 2.73 & & \textbf{2.67} & 416,744   & & 1.42 & & \textbf{0.66}&  416,864\\
			\bottomrule
		\end{tabular}
	\caption{MAE on unseen data and number of trainable parameters for different pairs for outer and inner graph mechanism for a \textbf{3-layer} architecture for the \textbf{4-node graph} and \textbf{10-node graph}}\label{model_large}
	\end{table}



	\section{Discussion}
	Position plays a crucial role for the informedness in a \ac{ncde}. The position of our informed \ac{ncde} approach can be seen analogous to a shortcut connection in a \ac{resnet} \cite{He_2015}, only between spatial nodes and not in the virtual time dimension. We suspect a greater impact on prediction performance with larger graphs with sparse connectivity structure. We also observe faster convergence during training for architectures with outer informedness. Informing at the inner and the outer position simultaneously with the same matrix seems to deteriorate performance. Introducing a fully-connected trainable outer matrix in an attempt to learn the adjacency matrix led to exploding gradients. In summary we demonstrated that prior graph information can benefit model performance of a \ac{ncde} on a physical time series forecasting task. The position in the architecture where one inserts the information is crucial. One has to investigate further if graph information is detrimental to learning non-causal temporal patterns across distant vertices.
 We dedicate our future research to exploring larger graphs, integrating graph edge features, and addressing non-constant advection, with the goal of applying our topology informed GNCDE to real-world data that includes exogenous influx at the vertices and probabilistic behavior.
	
	\section*{Acknowledgements}
	This work was supported by the Federal Ministry for Transportation and Digital Infrastructure (BMDV) as grant SOLP (19F2204B); and the Federal Ministry for Economic Affairs and Climate Action (BMWK) as grant DAKI-FWS (01MK21009A).
	

	
	\bibliography{d3s3_neurips_2024}

	\newpage
	\appendix
	\section{Appendix}
	
	\subsection{Dimensions}{\label{dimensions}}
	The dimensions of the domains and codomains of the  vector field functions $f_{\boldsymbol{\theta}}$ and $g_{\boldsymbol{\gamma}}$ can be described even more precisely with
	\begin{equation*}
		f_{\boldsymbol{\theta}}(\cdot) : \mathbb{R}^{|\mathcal{V}| \times d_h} \rightarrow \mathbb{R}^{|\mathcal{V}| \times d_h} \times \mathbb{R}^{|\mathcal{V}| \times d_x}
	\end{equation*}
	\begin{equation*}
	g_{\boldsymbol{\gamma}}(\cdot): \mathbb{R}^{|\mathcal{V}| \times d_z} \rightarrow \mathbb{R}^{|\mathcal{V}| \times d_z} \times \mathbb{R}^{|\mathcal{V}| \times d_h}
	\end{equation*}
	and $\bold{H}(t) \in \mathbb{R}^{|\mathcal{V}| \times d_h}$, $\bold{Z}(t) \in \mathbb{R}^{|\mathcal{V}| \times d_z}$, $\bold{X}(t) \in \mathbb{R}^{|\mathcal{V}| \times d_x}$, with $d_h$, $d_z$ and $d_x$ being the dimensionality of the corresponding (hidden) states or control per vertex,
	leading to a description of the tensor product in Equation (\ref{temporal}) with einstein notation while omitting the $\boldsymbol{\theta}$ and $\boldsymbol{\gamma}$ for more clarity
	\begin{align*}
	\left(\frac{d\bold{H}}{dt} \right )_{mh} = \left (f \frac{d\bold{X}}{dt} \right )_{mh} = f_{mhnx} \left (\frac{d\bold{X}}{dt}\right )_{nx}
	\end{align*}
	and for Equation (\ref{spatial})
	\begin{align}\label{tensor_multi_g}
	\left (\frac{d\bold{Z}}{dt} \right )_{kz} = \left (g \frac{d\bold{H}}{dt} \right )_{kz} = g_{kzmh} \left (\frac{d\bold{H}}{dt} \right )_{mh},
	\end{align}
	where $n,m$, and $k$ are indices for the vertices, $h$ and $z$ are the indices for the dimension of hidden states $\bold{H}$ and $\bold{Z}$, respectively and $x$ is the index for the dimension of the control path $\bold{X}$.
	
	If one now wants to incorporate graph information at the outer position in Equation (\ref{outer_info}), one would insert $\bold{A}_{\mathcal{E}}$ into Equation (\ref{tensor_multi_g}) and obtain
	\begin{align*}
		\left (\frac{d\bold{Z}}{dt} \right )_{kz} = \left (g \frac{d\bold{H}}{dt} \right )_{kz} = g_{kzmh} \ \left (\bold{A}_{\mathcal{V}} \right )_{mn} \ \left (\frac{d\bold{H}}{dt} \right )_{nh}.
	\end{align*}
	
	\subsection{Adjacency matrices}\label{adjacencies}
	The adjaceny matrices for the 4-node graph and the 10-node graph are given by
	\begin{align*}
		\bold{A}_{\mathcal{V}_4} = \begin{bmatrix}0&1&0&0\\
			0&0&0.3&0.7\\
			0&0&0&1\\
			1&0&0&0
		\end{bmatrix}, 
		\quad
		\bold{A}_{\mathcal{V}_{10}} =
		\begin{bmatrix}
			0 & 1 & 0 & \ldots & 0 \\
			0 & 0 & \ddots & \ddots & \vdots \\ 
			\vdots & \vdots & \ddots & 1 & 0\\
			0 & \ldots & 0 & 0 & 1 \\
			0 & \ldots & 0 & 0 & 0
		\end{bmatrix}.
	\end{align*}
	
	\subsection{Edge transition matrix}\label{edge_transition}
	To obtain the edge transition matrix $\bold{A}_{\mathcal{E}}$ from a weighted vertex adjacency matrix $\bold{A}_{\mathcal{E}}$ with $\sum_{j \in \vert \mathcal{E} \vert} (\bold{A}_{\mathcal{E}})_{ij} = 1$, one first starts off with the vertex incidence matrix $\bold{I}$. Let us define some matrix operations for $\bold{I} \in \mathbb{R}^{\mathcal{|V|} \times \mathcal{|E|}}$, namely
	\begin{align*}
		(\bold{I}^{+})_{ij} = \begin{cases}
			(\bold{I})_{ij}, \quad  &\text{if } (\bold{I})_{ij} > 0 \\
			 0, \quad &\text{else}
		\end{cases},
	\end{align*} 
	\begin{align*}
	(\bold{I}^{-})_{ij} = \begin{cases}
		-(\bold{I})_{ij}, \quad  &\text{if } (\bold{I})_{ij} < 0 \\
		0, \quad &\text{else}
	\end{cases},
	\end{align*}
	\begin{align*}
	(\bold{I}^{c})_{ij} = \begin{cases}
		1, \quad  &\text{if } (\bold{I})_{ij} > 0 \\
		(\bold{I})_{ij}, \quad &\text{else}
	\end{cases},
	\end{align*} 
thereby $(\cdot)^c$ is making $\bold{I}$ conservative. In total the edge transition matrix $\bold{I}_{\mathcal{E}}$ is given by
\begin{align*}
	\bold{A}_{\mathcal{E}} = (\bold{I}^{-})^\intercal (\bold{I}^{c})^{+}.
\end{align*}
On the one side $(\bold{I}^{-})^\intercal \in \mathbb{R}^{|\mathcal{E}| \times |\mathcal{V}|}$ considers the incoming magnitude of the transported quantities from edges to vertices and also the amount of quantities, on the other side $(\bold{I}^{c})^{+} \in \mathbb{R}^{|\mathcal{V}| \times |\mathcal{E}|}$ distributes the outgoing quantites from the vertices to the edges while solely encoding connectivity and neglecting magnitude.
	
\subsection{Applications}\label{applications}
For traffic forecasting, one would consider the traffic flow sensors the nodes of the graph  and the road network would be the edges. Analogously in the climate and weather forecasting scenario, weather observation stations measuring quantities like precipitation, temperature, air pressure and wind, would constitute the nodes, and the edges would transport abstract representations that encode the propagation of the physical quantities in space. Here the propagation in three-dimensional space would be encoded as an information link which is represented as a multivariate function on every one-dimensional domain on the edges. In river water level forecasting, one considers the gauging stations as nodes and directed edges that are derived from the geographical river topology connections.
For disease spread, a node would be regarded as an administration area for which one can measure the aggregate infections, and the edges would connect these regional areas according to mobility connections, e.g. adjacent areas would be connected, areas that can be reached via train or airplane etc. Here the velocity field on the edges would need to be adapted according to the speed of the means of travel.

	\subsection{Causality}
In total we moved from a space-continuous description of the state on the edges and vertices during the simulation to a discrete space on the vertices for the measurements. In the continuous advection description, we had a clear cause-effect relation due to the fact that we could look back along an edge or back to previous edges with the edge transition matrix. This continuous description of the causal structure is now subsumed to aggregated, discrete values at nodes and edge connections without an attached domain between them. To be able to incorporate this causal structure into the \acp{gncde} we have to consider a long enough context window to be able to receive information from the previous vertices at earlier times. The task then resembles learning a delay differential equation.
	\subsection{Additional Results}
	In Table \ref{4_node_model} the MAE on the 4-node graph for different \ac{gncde} configurations with only 2 layers is reported.
	\begin{table}[H]
		\caption{MAE on unseen data and number of trainable parameters for different selection of pairs for outer and inner graph mechanism for the smaller model with a \textbf{2-layer} architecture for the \textbf{4-node graph}}\label{4_node_model}
		\begin{tabular}{cccccccc}
			\toprule
			   & \multicolumn{5}{c}{$\bold{A}_{\mathcal{V}^{outer}}$} \\
			& \multicolumn{2}{c}{Identity} & & \multicolumn{2}{c}{Informed} \\\cmidrule{2-3} \cmidrule{5-6}
			$\bold{A}_{\mathcal{V}^{inner}}$ & MAE & \# params & & MAE & \# params \\
			\midrule
			Identity & 3.63 & 289,024 & & 2.96 & 289,024 \\ 
			Informed & 3.21 & 289,024 & & 3.60 & 289,024 \\
			AGC 	 & 2.83 & 412,584 & & \textbf{2.78} & 412,584 \\
			\bottomrule
		\end{tabular}	
	\end{table}

	In Table \ref{smalusepackagel_model_10} the MAE on the 10-node graph for different \ac{gncde} configurations with only 2 layers is reported. 
	
	\begin{table}[h]
		\caption{MAE on unseen data and number of trainable parameters for different selection of pairs for outer and inner graph mechanism for the smaller model with a \textbf{2-layer} architecture for the \textbf{10-node graph}}\label{smalusepackagel_model_10}
		\begin{tabular}{cccccccc}
			\toprule
			 & \multicolumn{5}{c}{$\bold{A}_{\mathcal{V}^{outer}}$} \\
			& \multicolumn{2}{c}{Identity} & & \multicolumn{2}{c}{Informed} \\ \cmidrule{2-3} \cmidrule{5-6}
			$\bold{A}_{\mathcal{V}^{inner}}$ & MAE & \# params & & MAE & \# params \\
			\midrule
			Identity & 2.80 & 289,084 & & \textbf{1.92} & 289,084 \\ 
			Informed & 2.62 & 289,084 & & 2.75 & 289,084 \\
			AGC 	 & 2.09 & 412,074 & & 1.94 & 412,072 \\
			\bottomrule
		\end{tabular}	
	\end{table}
\subsection{Background on Neural Differential Equations}\label{nde_background}
\textbf{Neural Ordinary Differential Equations.}
\acp{node} (\citet{chen_neural_2019}) are time-continuous models that use the structure of a differential equation to model the evolution of state variables $y \in \mathbb{R}^{d}$:
\begin{align*}
	y(0) = y_{0}, \quad \frac{\dif y}{\dif t} = f_{\theta}(t, y(t)).
\end{align*}
The Lipschitz-continuous function $f_{\theta}: \mathbb{R} \times \mathbb{R}^d \rightarrow \mathbb{R}^{d}$ on the right hand side, the so called vector field that guides the derivative in the equation is parametrized by an \ac{nn} with parameters $\theta$, and thereby rendering it "neural". \acp{node} can be considered the continuous-time limit of residual networks. 
One can bring an \ac{node} also into an integral formulation:
\begin{align*}
	y(0) = y_{0}, \quad y(t) = y(0) + \int_{0}^{t} f_{\theta}(s, y(s)) \dif s.
\end{align*}
\textbf{Neural Controlled Differential Equations.}
In comparison to the aforementioned \ac{node} formulation in integral form, in a \ac{ncde} \cite{kidger_neural_2020} described by
\begin{align*}
		y(0) = y_{0}, \quad y(t) = y(0) + \int_{0}^{t} f_{\theta}(y(s)) \dif x(s),
\end{align*}
the continuous path $y: [0,T] \rightarrow \mathbb{R}^{d_y}$ must satisfy the equation which is driven by a control path $x(s): [0,T] \rightarrow \mathbb{R}^{d_x}$ that allows continuously introducing information from data at later points in time. $d_x, d_y \in \mathbb{N}$ are the dimensionalities of the two paths $x$ and $y$, respectively. The vector field $f: \mathbb{R}^{d_y} \rightarrow \mathbb{R}^{d_y \times d_x}$ must be Lipschitz-continuous. \acp{ncde} can be considered the continuous version of \acp{rnn}.
\end{document}